\documentclass[11pt,a4paper]{article}
\usepackage[hyperref]{acl2020}
\usepackage{times}
\usepackage{enumitem}
\usepackage{graphicx}
\usepackage{scalerel,xparse}
\usepackage{latexsym}
\usepackage{url}
\usepackage{breakurl}
\usepackage{float}
\usepackage{tikzsymbols}

\NewDocumentCommand\thinking{}{
    \includegraphics[scale=0.05]{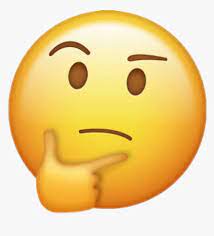}
}
\NewDocumentCommand\shock{}{
    \includegraphics[scale=0.05]{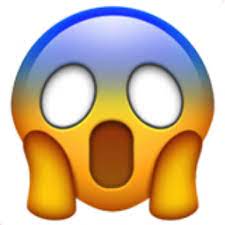}
}
\NewDocumentCommand\happy{}{
    \includegraphics[scale=0.05]{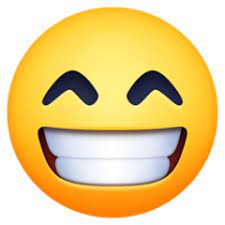}
}
\NewDocumentCommand\tiredcry{}{
    \includegraphics[scale=0.05]{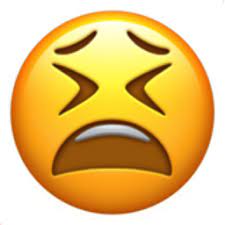}
}
\NewDocumentCommand\pray{}{
    \includegraphics[scale=0.05]{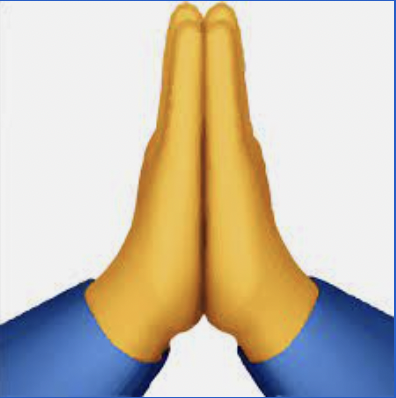}
}
\NewDocumentCommand\eyeroll{}{
    \includegraphics[scale=0.05]{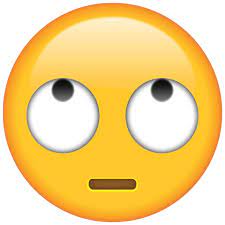}
}

\NewDocumentCommand\fire{}{
    \includegraphics[scale=0.05]{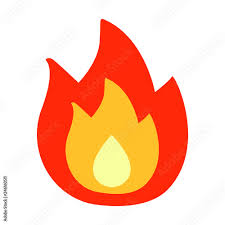}
}

\usepackage{microtype}

\aclfinalcopy 

\title{Token-free Models for Sarcasm Detection}

\author{Kanika Agarwal \\
  New York University \\
  \texttt{ka2522@nyu.edu} \\\And
  Maitreya Sonawane \\
  New York University \\
  \texttt{mss9240@nyu.edu} \\
  \And
  Sumit Mamtani \\
  New York University \\
  \texttt{sm9669@nyu.edu} \\
  \And
  Nishanth Sanjeev \\
  New York University \\
  \texttt{ns5287@nyu.edu} \\
}
\date{}

\begin{document}
\maketitle
\begin{abstract}

Tokenization has traditionally been a major step toward approaching a natural language processing (NLP) task. However, it comes with its own set of downsides. Recent developments show that operating on raw text in a character or byte-level fashion provides many benefits. Our study evaluates these 'token-free models' - specifically, ByT5 and CANINE, on social media tasks (particularly, sarcasm detection). We fine-tune and evaluate the performance of these models in a social media domain, i.e., Twitter tweets, and compare these results to their performance in a non-social media domain, i.e., news headlines. Our experiments show that token-free models perform better than their token-level counterparts in the social and non-social media domains. In a fair comparison, ByT5-small and CANINE outperform the current SotA approaches on the News Headlines and Twitter Sarcasm datasets. In terms of accuracy, ByT5-small and CANINE exceed these approaches by 0.77\% and 0.49\%, respectively. Therefore, analysing the performance of token-free models can further contribute to NLP research in social media tasks.

\end{abstract}



\section{Introduction}

Token-free models are a recent innovation in NLP that operate directly on raw text. This contrasts with traditional models that rely on processing textual data in a token-wise fashion.
Although NLP researchers generally use tokenization as a preprocessing procedure, there may be issues
caused due to the disadvantages of explicit tokenization. For instance, the presence of typos, as well as tokens that are out-of-vocabulary (OOV), can cause issues with text processing \citep{https://doi.org/10.48550/arxiv.2010.12663}. Additionally, most tokenizers are not language-agnostic, i.e., a tokenizer trained on text written in English, cannot be used to tokenize text in a non-English language \citep{https://doi.org/10.48550/arxiv.2012.15613}. Furthermore, traditional tokenizers may not account for the presence of certain non-word entities, such as emojis/emoticons. However, when approaching a sentiment analysis task (such as sarcasm detection), it is essential to account for the presence of such entities, as they may contribute valuable information regarding the underlying sentiment of a given statement.\\
\hspace*{5mm} Byte-level models, specifically ByT5 \citep{https://doi.org/10.48550/arxiv.2105.13626}, have proven to be competitive with their token-level and subword-level counterparts, respectively. Additionally, these models remedy some of the disadvantages of using traditional tokenizers. For instance, byte-level models can handle OOV words and are more robust to noise. Furthermore, these models can process text in any language \citep{https://doi.org/10.48550/arxiv.2105.13626}. Additionally, these models do account for the presence of entities like emojis and emoticons in text.\\
\hspace*{5mm} Now, sarcasm detection is a classification problem in NLP that finds applications in opinion mining
and marketing research. Considering that token-free models are in some ways superior to traditional
token-based models (as they overcome the above-stated disadvantages of explicit tokenization), we evaluate the performance of a token-free model on a sarcasm detection task with social media data (which is novel, according to our knowledge). Bearing in mind that token-free models can process text in any language and are robust to noise, we hypothesize that they perform well in a social media environment. \\
\hspace*{5mm} Furthermore, as byte-level models significantly perform better on tasks sensitive to spelling and
pronunciation, we believe that they can perform well on a sarcasm detection task in a social media environment. Prior literature indicates that non-linguistic cues, such as special keyword characters and emoticons, play an essential role in detecting sarcasm \cite{gonzalez-ibanez-etal-2011-identifying}. Most token-based models such as T5 and BERT do not understand emoticons, as their tokenizers haven't been trained on them \cite{DBLP:journals/corr/abs-1910-10683, DBLP:journals/corr/WuSCLNMKCGMKSJL16}. However, token-free models such as ByT5 and CANINE do not ignore emoticons. Since emoticons in a social media setting are vital for sarcasm detection, we believe that token-free models can perform well here. Furthermore, social media texts also contain multiple words joined together in the case of hashtags, which we expect can lead to token-based models failing because of incorrect tokenization. Previous studies \citep{https://doi.org/10.48550/arxiv.2107.02276} have classified research in sarcasm detection primarily into three paradigms: 1) semi-supervised pattern extraction to identify implicit sentiment, 2) use of hashtag-based supervision\cite{hastag-supervision}, and 3) incorporation of context beyond target text. We believe that token-free models will be able to assimilate the context in sarcastic text, and that we can train these token-free models on a supervised learning task, i.e., sarcasm detection, using annotated datasets from different domains. We elaborate on our experimental design and baseline result analyses in Section 3 below. \\
\hspace*{5mm} Finally, we compare the performance of the selected models on tasks from two different domains - social media and journalistic literature. We hypothesize that these models will experience a drop in accuracy while being tested on the former, as social media datasets contain noisy texts, including typos, emojis, grammatical errors, etc. The news dataset chosen contains sarcastic headlines that do not require hashtags-based supervision and do not require additional context, unlike tweets that might be replies to other tweets. Considering that token-free models can handle noise efficiently, we expect them to perform better than token-based models in the social media dataset as compared to sarcasm detection in the news dataset.\\
\hspace*{5mm} The contributions of our study are summarized as follows:
\begin{itemize}
    \item We propose that token-free models perform better than traditional token-based models for the sarcasm detection task. Specifically, we compare the performance of token-free models on datasets from two different domains - social media and journalistic literature.
    \item We provide a baseline for sarcasm detection in social media and journalistic literature. Our code and pre-trained models are available for further exploration in this domain.
    \item ByT5-small and CANINE-s achieve 89.87\% and 88.28\% on the News Headlines and Twitter Sarcasm datasets, respectively, thus outperforming the SotA approaches in a fair comparison.
\end{itemize}




\section{Background}
\label{sec:length}


\subsection{Automated sarcasm detection}

\hspace*{5mm} Early work on sarcasm detection depended on lexical \cite{kreuz-caucci-2007-lexical}, pragmatic (like emoticons or replies) \cite{davidov-etal-2010-semi}, punctuation \cite{tungthamthiti-etal-2014-recognition} and interjection word-based features \cite{ghosh-etal-2015-sarcastic}. Many supervised \cite{yeah-right}, semi-supervised \cite{tsur} and unsupervised \cite{nozza} techniques for automated sarcasm detection were explored based on SVM, Naive Bayes, and Random Forest classifiers using such textual features. Deep learning-based models, including RNNs and LSTMs \cite{Salim2020DeepLW} gained importance in this domain. As Twitter became a reliable source for data collection, the usage of hashtags for supervised sarcasm detection became prominent \cite{BHARTI2016108}. Furthermore, using contextual features beyond the target text has become popular for sarcasm detection, as the tone of the target text can depend a lot on the previous dialect.\\
\hspace*{5mm} \citet{gregory-etal-2020-transformer} implemented LSTM, GRU, and transformer-based models for sarcasm detection and found transformer-based models to be performing the best. 
Attention and transformer-based models have become popular for all NLP tasks in recent years. \\
\hspace*{5mm} \citet{DBLP:journals/corr/abs-2005-11424} used deep transformer layers using multi-headed attention on context along with target text. Most of the transformer-based models used for sarcasm detection are pre-trained on embeddings of tokens after lemmatization of input text. Few models have recently been used in literature for NLP tasks that do not rely on tokenizing the input text and instead, work directly at the character level. Token-free models have shown to perform better than token-based models for many NLP tasks (elaborated on in the next subsection).

\subsection{Token-free models in NLP}

\hspace*{5mm} \citet{DBLP:journals/corr/abs-2103-06874} proposed CANINE, which was the first pre-trained encoder using tokenization-free model. CANINE can operate directly on characters and combines downsampling with a deep transformer stack. The former reduces the input sequence length, while the latter encodes context. 
\\
\hspace*{5mm} \citet{https://doi.org/10.48550/arxiv.2105.13626} proposed ByT5, which is a variant of multilingual T5, i.e., mT5 \cite{DBLP:journals/corr/abs-2010-11934}, that can operate directly on a UTF-8 encoding of raw input text. The architecture of ByT5 is similar to that of T5, except the number of encoder layers is larger than the number of decoders. The authors evaluated these models on benchmarks and downstream tasks, including GLUE \cite{wang-etal-2018-glue}, SuperGLUE \cite{DBLP:journals/corr/abs-1905-00537}, XSum \cite{xsum-emnlp}, TweetQA \cite{DBLP:journals/corr/abs-1907-06292}, and DROP \cite{DBLP:journals/corr/abs-1903-00161}. ByT5 was shown to outperform T5 in scenarios sensitive to the presence of noise in text and on tasks sensitive to spelling.\\
\hspace*{5mm} \citet{DBLP:journals/corr/abs-2106-12672} proposed Charformer. Here, instead of passing encodings to single characters (like in CANINE or ByT5), Charformer uses character n-grams. It contains two parts - the first one learns representations of the input text from its character n-grams, and the second part is a model on top of it for scoring purposes. Since token-free models are computationally expensive, Charformer uses an end-to-end learning gradient-based subword tokenization (GBST). This brings together the efficiency of subword tokenization with the advantages of character-level representations. By scoring candidate subword blocks using a scoring network, GBST develops a position-wise soft selection over them. GBST learns interpretable latent subwords, allowing for simple lexical representation analysis. Charformer mostly outperforms subword-based models when run on GLUE, multilingual, and noisy text datasets.

\section{Experimental setup}
\hspace*{5mm} We validate our work using a sarcasm dataset of Twitter tweets, and the Kaggle News Headlines Dataset for Sarcasm Detection.
The News Headlines dataset is collected from two news websites -  TheOnion\footnote{\url{https://www.theonion.com/}} (sarcastic) and HuffPost\footnote{\url{https://www.huffpost.com/}}  (non-sarcastic). 
The Kaggle News dataset has some advantages over the existing Twitter datasets, such as having no spelling mistake/informal usage, hence, reducing sparsity. The News dataset can also be considered to be of higher quality sarcastic content as the sole purpose of TheOnion is to publish sarcastic news.

Here, each record contains three attributes - \textit{is\_sarcastic} (0/1 depending on sarcastic label), \textit{headline} (source\_text), and \textit{article\_link}. The size of this dataset is 6.06 MB and contains 28619 headlines, each with a label to manifest if it is sarcastic.
\\
\hspace*{5mm} Meanwhile, the Twitter sarcasm dataset contains two JSON files for training and testing. Each of these JSON files contains a list of dictionaries, each with 3 fields - \textit{context}, \textit{label}, and \textit{response}. The \textit{response} field contains the actual tweet text, while the \textit{context} field contains a list of sentences that represent the contextual tweets to that response. Additionally, the \textit{label} field indicates whether the response is sarcastic or not (0/1 field). While the training file is 3.69 MB in size and contains 5000 tweets with their contexts and labels, the testing file spans 1.26 MB with 1800 similar tweets.

The baseline model we chose is \textit{T5-base} \citep{https://doi.org/10.48550/arxiv.1910.10683}, where the model is pre-trained on a data-rich task before being fine-tuned on a downstream task. For token-free models, we used: 1) \textit{ByT5-small} and 2) \textit{ByT5-base} \citep{https://doi.org/10.48550/arxiv.2105.13626} which are tokenizer-free extensions of the mT5 model and operate directly on UTF-8 bytes, 3) \textit{CANINE-s} \citep{DBLP:journals/corr/abs-2103-06874}, a tokenization-free \& vocabulary-free neural encoder pre-trained with subword loss, that operates directly on character sequences. More information about model sizes and extra experiments, including Charformer, in \autoref{appen} and \autoref{appenb}. \\
To finetune each model in our dataset, we passed specific data through their respective tokenizers. Then, depending on whether the model was imported from HuggingFace transformers \footnote{\url{https://huggingface.co/docs/transformers/index}} or not, it had a specific model trainer and evaluate() function to run the finetuning process. The hyperparamters were set in this stage and those for the best model for each dataset are reported in \autoref{novel}.

\section{Novel Techniques and Results} \label{novel}

\subsection{Sarcasm Detection Twitter}
\hspace*{5mm} As part of \textit{The Second Workshop on Figurative Language Processing}, \citet{https://doi.org/10.48550/arxiv.2005.05814} reported F1-scores by several competitors. At the end of the competition, HuggingFace released a T5 model finetuned on the same dataset getting the same F1 score as the benchmark. As the participants did not release their code publicly, we used the HuggingFace model as our benchmark. We evaluated it on the Sarcasm Detection Twitter dataset. 

To finetune CANINE-s on the Twitter dataset, we trained it for 15 epochs. We noticed that increasing learning rate decreased the accuracy, so we chose $lr=2e-5$. We set the $batch\_size$ for both train and eval sets to $16$. Similarly, we finetuned the rest of the models and the test accuracy we achieved is shown in \autoref{table:main-table}.

\subsection{News Headline dataset}
\hspace*{5mm} This is a Kaggle dataset worked on by several participants. The most upvoted notebook\footnote{\url{https://www.kaggle.com/code/madz2000/sarcasm-detection-with-glove-word2vec-83-accuracy}} has achieved 83\% accuracy using GloVe and Word2Vec models. On analyzing the other notebooks, we found that the accuracy was similar. To achieve a concrete baseline, we finetuned the T5 model on this dataset with 60\% data used for training, 20\% for validation, and 20\% for testing. In a similar manner with the previous dataset, we finetuned ByT5-small, ByT5-base, and CANINE-s on this dataset too. To finetune ByT5-small, we found the best hyperparameters (after running experiments) to be  $lr=0.01$, $batch\_size=8$ for both the train and eval set, $source\_max\_token\_len=1024$, $ target\_max\_token\_len = 16$. We trained it for 10 epochs. The results are as follows:

\begin{table}[h]
\scriptsize
\centering
\begin{tabular}{c|c|c|c|c}
\textbf{Dataset} & \textbf{T5-base} & \textbf{ByT5-small} & \textbf{ByT5-base} & \textbf{CANINE-s} \\ \hline
\textit{Twitter} & 72.39\%          & 65.73\%             & 69.22\%            & \textbf{72.88}\%           \\
\textit{News}    & 89.10\%          & \textbf{89.87}\%             & 85.59\%            & 88.28\%          
\end{tabular}
\caption{The table shows comparison of performance of various models on Sarcasm datasets from two different domains}
\label{table:main-table}
\end{table}

\section{Analysis}
\subsection{Twitter Dataset}
\hspace*{5mm} We found that CANINE-s outperformed the baseline model by 0.49\%. We attribute this to ability of CANINE to combine downsampling, which reduces the input sequence length that can encode the tweet context better. 
\subsection{News Dataset}
\hspace*{5mm}  Here, the highest accuracy was achieved by ByT5-small and surpassed its baseline by 0.77\%. Given the capability of ByT5 to work best on short-to-medium length text sequences, and headlines being shorter than context-incorporated tweets, ByT5 performs the best.
\subsection{Model Comparison}
\hspace*{5mm}  On investigating outputs from the two types of models on the Twitter dataset, we found that token-free models are able to correctly identify the presence or absence of derision where token-based models failed to do so. Most of these examples contained emojis, spelling errors, abbreviations, or slang terms. Even though we expected token-free models to perform better in the presence of hashtags, we found both types of models to be performing quite similarly. This is because the tokenizers of token-based models can properly separate mixed words into subwords. Although the Twitter dataset is mainly in English, we found a few statements written in Hindi. For these examples, we found T5 to fail, whereas ByT5 (an extension of mT5) correctly infers sarcasm. Furthermore, we observe ByT5-small outperforming T5 on the News dataset but not on the Twitter dataset. More analysis about examples in the \autoref{appen}. \\
\hspace*{5mm} Finally, on comparing results in a social media vs non-social media domain, we find token-free models to outperform token-based models. We expected token-free models to perform substantially well in social media domain. However, while they did score higher than the baseline, our token-free models couldn't outperform as much as they did on the News dataset. The differences are likely due to the computational limit we had, as we needed to truncate the maximum length of sequences from Twitter dataset and thereby lost valuable information. Another reason could be the size of Twitter training dataset, which may be too small to train a model as large as ByT5.

\section{Conclusion}
\hspace*{5mm} We compare the performance of token-free models on datasets from two domains - social media and journalistic literature. We find that token-free models outperform traditional token-based models in sarcasm detection tasks in both domains. CANINE and ByT5 outperform every other model, including the baseline T5, in both Twitter and News Headlines datasets. Finally, we scrutinize where the token-based models fail while token-free models produce accurate results. Overall, we believe that the results presented in our study facilitate the use of token-free models for other social media tasks in the future.


\section{Ethical Consideration}
Our study analyzed the performance of token-free models on sarcasm detection for social media tasks. While the Twitter sarcasm dataset is publicly available, it does not contain any user handles of those who tweeted. Any user handles that were mentioned in the tweets were replaced with an '@USER' placeholder. Furthermore, the News Headlines dataset used only contains headlines with no description of any specific individual from the news body. We expect future work on social media tasks will follow similar constraints when gathering user/public data. \\
\hspace*{5mm} Positive aspects of our study include attempting to understand the true sentiment of the person behind the screen (who shares information on social media) - which would, in turn, create a more intelligent AI to make Human-Computer interaction more natural, for instance. However, possible adverse effects to keep in mind include intrusive AI surveillance on social media platforms by companies.

\section{Collaboration statement}

All four of us played an equal part in conducting the literature survey, writing the partial draft paper, and generating baseline results. Specifically:\\

\textbf{Kanika Agarwal} worked on fine-tuning ByT5-base on the News dataset, contributed to testing T5 and ByT5-small on the Twitter data, fine-tuning CANINE on the Twitter data, investigated outputs from different models to do the analysis, and wrote Section 2.1, 2.2 and 5.3.\\

\textbf{Maitreya Sonawane} contributed to testing T5 on the Twitter sarcasm dataset, finetuning and testing T5 on News Sarcasm dataset, finetuning and testing CANINE on Twitter Sarcasm dataset and News Sarcasm dataset, Section 1, 4, 5 and 7 of the paper. \\

\textbf{Nishanth Sanjeev} wrote the Abstract, Introduction (Section 1), Section 2.2, prepared the Twitter sarcasm dataset for processing, finetuning and testing T5 and ByT5-small on the Twitter data.\\

\textbf{Sumit Mamtani} wrote and tested the ByT5-small code and Charformer model code on the Kaggle News sarcasm dataset and the Twitter sarcasm dataset. Contributed and tested the code for the CANINE model on the Twitter sarcasm dataset. Also wrote Section 3 (Experimental setup) in the Paper.

\section{GitHub}
All code files for the current project can be found
here:
\url{https://github.com/sonawanemaitreya/NLU_S22_Project}

\section{Acknowledgment}

\appendix

\section{Appendices} \label{appen}
\label{sec:appendix}
\begin{table}[H]
\scriptsize
\centering
\begin{tabular}{|c|c|c|c|c|}
\hline
\textbf{}                 & \textbf{T5-base} & \textbf{ByT5-small} & \textbf{ByT5-base} & \textbf{CANINE-s} \\ \hline
\textit{\# of parameters} & 220 M            & 300 M               & 580 M              & 121 M             \\ \hline
\end{tabular}
\caption{The table shows number of parameters in each model used in experimentation}
\label{table:size-table}
\end{table}
For Twitter dataset, few examples from test set where token-free models could correctly classify the input unlike T5 are given below:\\
\fbox{\begin{minipage}[c]{20em}
Text: U . K . faces ‘ major ’ coronavirus outbreak , world expert warns . A fourth patient in England has been diagnosed with coronavirus as a microbiologist warned that the country could suffer a “ major outbreak ” which is likely to become a pandemic $<$URL$>$@USER Not to worry . * 45 assures us it will just go away in April . \shock $<$URL$>$@USER @USER I feel so much better now\\
Analysis: Presence of emojis (\shock).\\
 \end{minipage}}
 \\
 \fbox{\begin{minipage}[c]{20em}
Text: It's a lazy \#SuperBowl Sunday pregame edition of \#AskBen , your questions and my answers until it gets repetitious and stale@USER Hey Ben ! When I listen to the THE FIFTH HOUR of your PODCAST your voice sound different ! Do you use a microphone from another dimension when you ’ re recording ? \thinking \#AskBen \happy $<$URL$>$@USER @USER I was thinking the same thing .@USER @USER @USER He ’ s badly overmodulating . Bad job by you Ben@USER @USER @USER That's because doesn't know how to produce audio levels .@USER @USER @USER don ’ t blame me for your buckled voice . Get out of bed at an earlier time . \#SuperBowl\\
Analysis: Presence of emojis (\thinking, \happy) and hashtags (\#SuperBowl).\\
 \end{minipage}}
\\
 \fbox{\begin{minipage}[c]{20em}
Text: \#day2 \#quaratine in \#diamondprincess 71 test done so far w / addition 10 positive result that accumulated 20 infected . I am getting worried and afraid . There is more test result to come . \tiredcry \tiredcry @USER Take care ! Relax and adequate exercises help a lot . Be optimistic ! Thank God you all are now in Japan . Imagine what if you were in China . How will you be treated there ? So many Chinese are now dying like dogs at their houses !  \pray@USER Why you saying this . Virus attacks any races . Nobody deserves it not a right time to called people die like dog \eyeroll \eyeroll \eyeroll my Twitter is focus on spreading \#PositiveEnergy .\\
Analysis: Presence of emojis (\tiredcry, \pray, \eyeroll) and hashtags.\\
------------------------------------------------------------

Text: While the calls for genuine access to be granted to students are legitimate , we must be very alarmed at the use of violence to make the point . Arson is not a legitimate way to request access . Students undermine the genuine call for corridors of learning to be open by doing thisTwo problems are created instead of one . Firstly the problem of funding continues to exist . Secondly a new funding challenge is created to replace the damaged infrastructure and furniture . University administrators then feel a need to get extra security@USER These are the ppl that will tomorrow be out there looking for a job . You give them a job and your business goes in \fire \fire \fire . The stupid mentality of my local ppl to always destroy property must stop . Tomorrow they will go burn a clinic demanding that govt fix UKZN .	\\
Analysis: Presence of emojis (\fire) and abbreviations (`\textit{govt}', `\textit{ppl}')\\
------------------------------------------------------------

Text: HOW MANY ARE GOING TO ST IVES ? ? So you are allowed to bring your family members into d hallowed chamber as it is called ? Asking for a friend .. 4 wives , 27 children and counting ... What bills will this person be supporting except the ones that put more cash in his pocket ...@USER Like I always tell people followership are not the problem but leadership ... now I see why this country is not moving forward .@USER @USER Very wrong it has always been the follwers ... we condone all thei excesses so why they misbehave and we look the other way and say what is my business putting the future generation in a bigger problem than the one we are already in ...
\par Analysis: Presence of spelling errors (`\textit{thei}', `\textit{follwers}')
\end{minipage}}

\section{Charformer} \label{appenb}

\begin{table}[h]
\scriptsize
\centering
\begin{tabular}{c|c}
\textbf{Dataset} & \textbf{Charformer} \\ \hline
\textit{Twitter} & $\sim$ 59\% \\                     \\
\textit{News}    & $77.6\%$ \\
\end{tabular}
\caption{The table shows comparison of performance of charformer model on Sarcasm datasets from two different domains }
\label{table:main-table}
\end{table}
We trained Charformer from scratch, as the pre-trained model weights were not provided by the authors. We then tested it on both datasets. This gave us reasonably good results but could not outperform our baseline (T5-base). Therefore, we chose not to report it in our final draft. 
\end{document}